# Mining Complex Hydrobiological Data with Galois Lattices

A. Bertaux[1,2], A. Braud[2], F. Le Ber[1,3]

[1]CEVH UMR MA 101 - ENGEES
1 quai Koch, BP 61039, F 67070 Strasbourg cedex
{aurelie.bertaux, florence.leber}@engees.u-strasbg.fr

[2]LSIIT UMR 7005
Bd Sébastien Brant, BP 10413, F 67412 Illkirch cedex
agnes.braud@urs.u-strasbg.fr

[3]LORIA UMR 7503
BP 35, F 54506 Vandœuvre-lès-Nancy cedex

## Abstract

*We have used Galois lattices for mining hydrobiological data. These data are about macrophytes, that are macroscopic plants living in water bodies. These plants are characterized by several biological traits, that own several modalities. Our aim is to cluster the plants according to their common traits and modalities and to find out the relations between traits. Galois lattices are efficient methods for such an aim, but apply on binary data. In this article, we detail a few approaches we used to transform complex hydrobiological data into binary data and compare the first results obtained thanks to Galois lattices.*

## 1. Introduction

Water quality is a major problem in Europe, underlined by the recent European Water Framework Directive. To evaluate the physico-chemical quality of a water body appeared to be not sufficient, new tools are required for evaluating the quality of the whole ecosystem [1]. Furthermore a comparison of existing tools and approaches is necessary to get a coherent monitoring of water bodies in Europe.

There exist several biological indices based on the faunistic and floristic species living in fresh water (e.g. five indices are used in France for qualifying running waters). These indices are useful, but it is difficult to compare their results from different areas, since the kind of species living in a river also depend on regional characteristics. A promising approach to avoid this drawback is to determine *functional* traits, shared by different species of different areas, that can be used to characterize water quality [8] or other ecosystems [7]. Currently, these functional traits have still to be defined for most of the categories of aquatic living species.

In this project, we focus on biological traits of european macrophytes, collected from the litterature, and we try to explore these data with Galois lattices [2,4]. Our aim is to find out sets of biological traits and species which can be interpreted as *functional groups* by the hydrobiologists.

The paper is organized as follows. First part is the current introduction, second part introduces the data, third part presents the methods we used to convert the data into a suitable format and the results we obtained with Galois lattices. The fourth part is a discussion on related work while fifth part gives some conclusions and perspectives of our work.

## 2. Biological traits of macrophytes

The data we deal with are about macrophytes (macroscopic plants living in water bodies, e.g. water lily). Each plant is described by a set of *traits* –or attributes– like potential size, reproduction period or anchorage mode... For each attribute there are several qualitative modalities. For example, the 'potential size' attribute owns four modalities: "under 0,08 meter", "between 0,08 and 0,3 meter", "between 0,3 and 1 meter", "between 1 and 5 meters". The 'reproduction period' attribute owns eight modalities (months from march to october)...

The modalities are associated to a value between 0 and 3 to indicate the *affinity* of the plants toward the modality. 0 means there is no plant having this modality, 1 means that a few plants have it, 2 a bit more, and 3 many. For example, the 'potential size' of *Berula erecta* (BERE) is given by the 4-set (1, 2, 3, 0) while it is (0, 1, 2, 2) for *Callitriche obtusangula* (CALO)*,* which means, in particular, that you will never find a *berula erecta* plant greater than 1 meter and no c*allitriche obtusangula* plant smaller than 0,08 meter (see Table 1).

The triple (trait, modality, affinity) allows to describe the biological characteristics of macrophytes in a qualitative and rather complex way.

For example, the data we deal with represent about 50 plants, described by 15 traits and 60 modalities. So, tools are needed to explore these data, and especially to cluster the plants according to their common traits and modalities and to find out the relations between various traits and modalities.

**Table 1. Traits data (potential size)**

| ATTRIBUTES | Potential Size | | | |
|---|---|---|---|---|
| | 1 | | | |
| MODALITIES | <0.08m | 0.08-0.3 m | 0.3 -1m | 1-5 m |
| AFFINITIES | 1 | 2 | 3 | 4 |
| BERE | 1 | 2 | 3 | 0 |
| CALO | 0 | 1 | 2 | 2 |
| ELOC | 0 | 2 | 3 | 1 |
| ELOE | 0 | 2 | 3 | 1 |
| ELON | 0 | 2 | 3 | 1 |
| LEMM | 3 | 0 | 0 | 0 |
| MENA | 0 | 1 | 3 | 1 |
| MYRS | 0 | 2 | 2 | 2 |
| NASO | 0 | 2 | 2 | 0 |
| NUPL | 0 | 0 | 1 | 3 |
| PTCO | 0 | 0 | 3 | 0 |
| PTNO | 0 | 0 | 2 | 3 |
| PTPE | 0 | 0 | 1 | 3 |
| RANU | 0 | 1 | 2 | 3 |
| SEFC | 0 | 1 | 3 | 1 |

## 3. Using Galois lattices on biological traits

Galois lattices are able to perform clustering on binary data and to extract implications sets of attributes [2,4,5]. Furthermore, work was done to adapt Galois lattices on more complex data [10].

In a preliminary step of our work, we decided to study the possibilities for using classical algorithms to build Galois lattices before using more complex and costly techniques. As those classical algorithms apply on binary data, we had to transform the original traits data. The transformations we used and the results we obtained -implication sets- are detailed hereafter. We worked with a subset of 15 plants described by 15 traits and 60 modalities.

Before, let us recall some definitions. Let *E* and *F* be two finite sets and *R* a binary relation on *E x F*. *E* is a set of objects, *F* a set of properties, *xRy* means that the object *x owns the property y*. Let *f* be a mapping from $2^E$ to $2^F$ such that, if *X* is an arbitrary part of $2^E$,

*f(X) = {y in F | for all x in X: xRy}*.

The mapping *g* is defined dually from $2^F$ to $2^E$ such that, if *Y* is an arbitrary part of $2^F$,

*g(Y) = {x in E | for all y in Y: xRy}*

The couple *{f,g}* is said to be a *Galois connection* between the sets *E* and *F*. From this connection, we get a set of concepts *(X, Y)*, such that *gof(X) = X* and *Y = f(X)*, that are organized within a lattice. *Y* is a set of attributes, called *intension*, and *X* is a set of objects, called *extension*. Furthemore, the lattice order allows to detect implication sets of properties and association rules [9].

## 3.1. Complete disjunctive table

Considering the original three levels format of the dataset, we transform it within a complete disjunctive table (or binary table) (Table 2). We denote the new attributes following a 'Lxx' model. The letter 'L' denotes a trait ('S' for potential Size, 'R' for potential of Regeneration...). The first 'x' is a number which indicates a modality and the second 'x' gives an affinity. For example, S21 means *"few plants (1) having a potential size (S) between 0,08 and 0,3 m ($2^{nd}$ modality)"*. For clarity purpose, we call those new attributes "properties" in the following.

The Galois lattice based on the disjunctive table is shown on Figure 1 (we show a sublattice including three traits, potential size, perennation and potential of regeneration). The whole lattice contains 1401 concepts, i.e. sets of macrophytes sharing the same modalities of the same traits with the same affinity. We have used the ConExp tool (for Concept Explorer [11]) both to build and to analyze the lattice. Actually ConExp allows to edit a context, to draw the associated lattice, to calculate the Duquenne-Guigues-Basis for implications between attributes, and to give the association rules that are true in this context.

**Table 2. The complete disjunctive table of traits data (potential size)**

|      | S10 | S11 | S13 | S20 | S21 | S22 | S30 | S31 | S32 | S33 | S40 | S41 | S42 | S43 |
|------|-----|-----|-----|-----|-----|-----|-----|-----|-----|-----|-----|-----|-----|-----|
| BERE |     | 1   |     |     |     | 1   |     |     |     | 1   | 1   |     |     |     |
| CALO | 1   |     |     |     | 1   |     |     | 1   |     |     |     |     | 1   |     |
| ELOC | 1   |     |     |     |     | 1   |     |     |     | 1   |     | 1   |     |     |
| ELOE | 1   |     |     |     |     | 1   |     |     |     | 1   |     | 1   |     |     |
| ELON | 1   |     |     |     |     | 1   |     |     |     | 1   |     | 1   |     |     |
| LEMM |     |     | 1   | 1   |     |     | 1   |     |     |     |     | 1   |     |     |
| MENA | 1   |     |     |     | 1   |     |     |     |     | 1   |     | 1   |     |     |
| MYRS | 1   |     |     |     |     | 1   |     |     | 1   |     |     |     | 1   |     |
| NASO | 1   |     |     |     |     | 1   |     |     | 1   |     | 1   |     |     |     |
| NUPL | 1   |     |     | 1   |     |     |     | 1   |     |     |     |     |     | 1   |
| PTCO | 1   |     |     | 1   |     |     |     |     |     | 1   | 1   |     |     |     |
| PTNO | 1   |     |     | 1   |     |     |     |     | 1   |     |     |     |     | 1   |
| PTPE | 1   |     |     | 1   |     |     |     | 1   |     |     |     |     |     | 1   |
| RANU | 1   |     |     |     | 1   |     |     |     | 1   |     |     |     |     | 1   |
| SEFC | 1   |     |     |     | 1   |     |     |     |     | 1   |     | 1   |     |     |

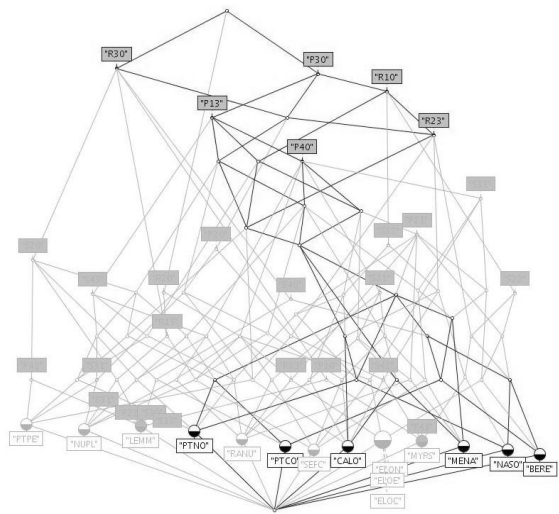

**Figure 1. The Galois lattice built from three traits of the complete disjunctive table**

The information provided by the lattice structure is interesting for hydrobiologists since they want to define *equivalences* between species with regard to their biological traits. For example, the Galois lattice in Figure 1 points out that the three plants ELON *(Elodea nuttallii),* ELOE *(Elodea ernstae)* and ELOC *(Elodea canadensis)* are grouped in the same concept –at the bottom of the lattice–with the following characteristics: P13, P21, P30, P40, S10, S22, S33, S41, R10, R20, R33. Actually, this can be directly read in the original table.

The concepts in the middle of the lattice are more interesting. For example, the concept highlighted in Figure 1 is the following : ((R10, R23, R30, P13, P30, P40) (PTNO, PTCO, CALO, MENA, NASO, BERE)). This concept means that the 6 (among 15) following plants, *Potamogeton nodosus, Potamogeton coloratus, Callitriche obtusangula, Mentha aquatica, Nasturtium officinale,* and *Berula erecta,* share the same following traits: potential of regeneration (low = 0, intermediate = 3, high = 0) and perennation (perennial underground organs = 3, bisannual = 0, annual = 0).

Furthermore we can extract the implication sets, and analyze them, e.g. *P13 => P30* (true for 13 individuals); *R23 => P30 R10 R30* (true for 8 individuals); for a better interpretation of the concepts. Finally we can say that the characteristics of these 6 plants are an intermediate potential of regeneration and perennial underground organs. This relationship between the two traits has still to be interpreted by hydrobiologists.

Considering the whole lattice (149 Lxx properties, 1401 concepts), we can extract 430 implication sets. 28 have a support equal to 14, 140 have a support between 5 and 9, and 262 between 1 and 4. Thus we obtain a few representative implications between traits.

Let us illustrate this with one of the implication sets which support is 14: *F10 A20 M10 => D30*. This rule means: F1≠0 or A2≠0 or M1≠0 or D30. For hydrobiologists it means that 14 plants have not a weak potential of dispersion (D30) or have a flexibility <10° (F1≠0) or a contact to the ground (A2≠0) or a reproduction period in march (M1≠0). Actually, looking at the original table you see that none of the 15 species have a reproduction period in march nor a contact to the ground. So, the final interpretation will be that all species (except one, *Nuphar lutea*) have a flexibility (> 10°) and an intermediate or high potential of dispersion. The implication set highlights the *mechanical* link between flexibility and dispersion.

Nevertheless, the conversion of the original data within a disjunctive table has three main problems. First, 1401 concepts give a lattice too huge to be readable. Second, the number of extracted implications is high. Third, it breaks an information which is meaningful for hydrobiologists, namely the distribution of the affinities of a macrophyte among the different modalities of a trait. We tried another approach to overcome this problem and present it in the following section.

### 3.2. Pattern approach

Before describing the new approach proposed, let us examine an illustrative example of the information we would like to represent. For instance, consider the plant BERE (*Berula erecta*), whose potential size is as follows (1, 2, 3, 0) according to the four modalities of this trait. This pattern (1, 2, 3, 0) is interesting for the hydrobiologists, because it shows the continuity of the size distribution of *Berula erecta*. Actually, having two plants with (almost) the same distribution is more meaningful than having two plants with the same affinity for one modality.

Thus, we have tried another conversion of the initial dataset. We have proposed to represent the distribution of the affinities of a plant according to the different modalities of a trait as a unique property, called a *pattern*. This pattern is composed as follows: first comes a letter that refers to the trait (like 'S' for potential Size) and then *n* numbers that refer to the affinity value of the modalities. For example S0122 means *"the potential size of members of this species is never of the first class (<0,08 m), sometimes of the second class (between 0,08 and 0,3 m), often of the third and fourth classes (between 0,3 and 1 m and between 1 and 5 m)"*.

The corresponding binary table -manually built- is shown on Table 3 for the potential size. Looking at this table, one can see that very few patterns are common to more than two individuals. The lattice built from these data has 76 concepts spread on 6 levels (excepting top and bottom). The lattice built for the three traits potential size, perennation and potential of regeneration, is shown on Figure 2. We can see that most of the patterns belong to only one individual.

**Table 3. Pattern table of traits data (potential size)**

| | S0013 | S0023 | S0030 | S0122 | S0123 | S0130 | S0131 | S0220 | S0222 | S0230 | S0231 | S1230 | S3000 |
|---|---|---|---|---|---|---|---|---|---|---|---|---|---|
| BERE | | | | | | | | | | | | 1 | |
| CALO | | | | 1 | | | | | | | | | |
| CHAR | | | | | | | | | | 1 | | | |
| CHAH | | | | | | 1 | | | | | | | |
| CHAV | | | | | | | | | 1 | | | | |
| ELOC | | | | | | | | | | | 1 | | |
| ELOE | | | | | | | | | | | 1 | | |
| ELON | | | | | | | | | | | 1 | | |
| LEMM | | | | | | | | | | | | | 1 |
| LEMT | | | | | | | | | | | | | 1 |
| MENA | | | | | | | 1 | | | | | | |
| MYRS | | | | | | | | | 1 | | | | |
| NASO | | | | | | | | | | 1 | | | |
| NUPL | 1 | | | | | | | | | | | | |
| NYMA | | | | 1 | | | | | | | | | |
| PTCO | | | | 1 | | | | | | | | | |
| PTNO | | | 1 | | | | | | | | | | |
| PTPE | 1 | | | | | | | | | | | | |
| RANC | | | | | 1 | | | | | | | | |
| RANU | | | | | | 1 | | | | | | | |
| SEFC | | | | | | | 1 | | | | | | |

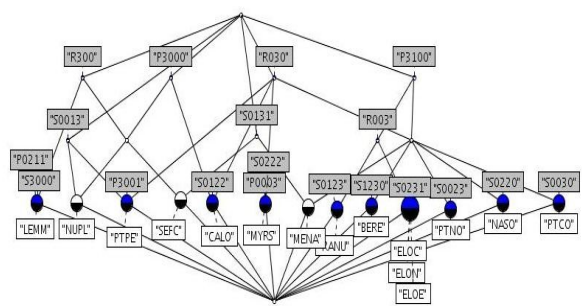

**Figure 2. The Galois lattice built from three traits of the pattern table**

Furthermore, from the whole lattice, 219 implication sets were extracted with a support under 5. This means only 5 plants (for the best result) support these implications. This is due to the patterns which are very precise and so few macrophytes match each of them. To solve this problem we can decrease the precision of the pattern, which can be done simply by grouping affinities. Either we consider the presence (affinities 1, 2 and 3 grouped together) and the lack (the affinity 0) of the modality, or we consider the affinity as low (affinities 0 and 1 grouped together) or high (affinities 2 and 3 gathered together).

The implications extracted following those methods have a support until 7 for the first solution and 8 for the second, which is much better. Nevertheless gathering those affinities is not pertinent for the hydrobiologists.

## 4. Discussion

The two approaches we studied until now are not very efficient according to the hydrobiologists requirement. The first one gives too much, unstructured information, while the second one gives very few but structured information. To explore further this second approach we will rely on [10] which proposed methods to deal with complex data within the Galois lattice theory. Actually [10] proposes to build and compare two lattices :

- Union lattice : the concept intent contains all the properties of the individuals belonging to the extent.
- Intersection lattice : the concept intent contains the properties belonging to all the individuals of the extent.

These lattices are built on specific Galois connections, depending on the object types (histogram, interval ...).

As our application deals with histogram data, $\Theta(x) = [\Theta^1, \Theta^2, \Theta^3 ..]$, we could use the following Galois connections.

Union:

$f(X) = [max_{x\ inX}\ \Theta^1, max_{x\ inX}\ \Theta^2, max_{x\ inX}\ \Theta^3 ...]$

$g(Y) = \{x\ |\ for\ all\ y\ in\ Y, \Theta(x) \leq y\}$

Intersection:

$f(X) = [min_{x\ inX}\ \Theta^1, min_{x\ inX}\ \Theta^2, min_{x\ inX}\ \Theta^3 ...]$

$g(Y) = \{x\ |\ for\ all\ y\ in\ Y, \Theta(x) \geq y\}$

This approach allows to compare two species for which trait pattern are different. For example, considering the two species *Berula erecta* and *Callitriche obtusangula* which size patterns are respectively [1, 2, 3, 0] and [0, 1, 2, 2], they could form a union-concept where the intent is [1, 2, 3, 2], and a intersection-concept where the intent is [0, 1, 2, 0]. In the ordinary way, there is no common size property between the two species (see Figure 2).

Other approaches are able to deal with such complex data, by building several lattices and then combining them, or by *cutting* big lattices (see e.g. [6]). Using fuzzy lattices [3] is another interesting way, since the affinity properties are very similar to probabilities.

## 5. Conclusions

Our aim is to help hydrobiologists in defining a new evaluation system of the quality of water bodies. In this paper, the main concern with respect to that problem is to extract knowledge from data that do not depend on regional characteristics. This is an important problem in order to be able to compare the quality of water bodies in different regions and to build a coherent evaluation system over Europe. Analyzing biological traits and determining functional groups is a promising approach for such an aim as they allow to evaluate water quality in more general way than the species themselves.

In this paper, we focus on the analysis of biological traits of macrophytes. In order to determine functional groups of macrophytes, we have proposed to use Galois lattices and have tried to extract groups of biological traits shared by groups of species, and to analyze implications between biological traits.

We have pointed out the fact that traits data are represented as triples (trait, modality, affinity) which make them too complex to directly build a lattice from them. We have thus proposed two conversions from those data to binary ones: building a full disjunctive table and using patterns which represent the distributions of species affinities *wrt* the modalities of biological traits. None of these approaches is really satisfactory. The first one gives too much, unstructured information, while the second one gives very few but structured information.

As further research, we propose to investigate the benefits of using lattices with a more complex structure, as those defined in [10]. Those lattices will allow us to extend the second approach studied, by building more general and thus more representative concepts. They should overcome the problems of both the approaches already used: the information on the distributions will be kept, but will be more general so that it will enable to extract more useful concepts. Furthermore, we want to explore the association rules provided by these lattices. To validate the approach, the concepts and rules extracted will be shown to the experts who have to give them a functional interpretation *wrt* aquatic ecosystems.

## 6. Acknowledgments

A. Bertaux and F. Le Ber thank the AERM, Agence de l'Eau Rhin-Meuse, for supporting this project.